\documentclass[sigconf]{acmart}

\usepackage{graphicx}
\usepackage{caption}
\usepackage{subcaption}
\usepackage{multirow}

\usepackage[linesnumbered,ruled,vlined]{algorithm2e}
\usepackage{ulem}
\AtBeginDocument{%
  \providecommand\BibTeX{{%
    \normalfont B\kern-0.5em{\scshape i\kern-0.25em b}\kern-0.8em\TeX}}}

\setcopyright{acmcopyright}
\copyrightyear{2018}
\acmYear{2018}
\acmDOI{10.1145/1122445.1122456}

\acmConference[Woodstock '18]{Woodstock '18: ACM Symposium on Neural
  Gaze Detection}{June 03--05, 2018}{Woodstock, NY}
\acmBooktitle{Woodstock '18: ACM Symposium on Neural Gaze Detection,
  June 03--05, 2018, Woodstock, NY}
\acmPrice{15.00}
\acmISBN{978-1-4503-XXXX-X/18/06}

\begin{document}

\title{Two-stage Voice Application Recommender System for Unhandled Utterances in Intelligent Personal Assistant}

\author{Wei Xiao$^{1, *}$, Qian Hu$^{2}$, Thahir Mohamed$^{2}$, Zheng Gao$^{2}$, Xibin Gao$^{2}$, Radhika Arava$^{2}$, Mohamed AbdelHady$^{2}$ \\
	$^{1}$AWS AI \, $^{2}$Amazon Alexa AI \\
	\{weixiaow, huqia, thahirm, zhenggao, gxibin, aravar, mbdeamz\}@amazon.com}
\thanks{* Work done while at Amazon Alexa AI}
%
%
%
%
%
%

\renewcommand{\shortauthors}{Wei et al.}

\settopmatter{printacmref=false} 
\renewcommand\footnotetextcopyrightpermission[1]{} 
\pagestyle{plain} 

\begin{abstract}
 Intelligent personal assistants (IPA) enable voice applications that facilitate people's daily tasks. However, due to the complexity and ambiguity of voice requests, some requests may not be handled properly by the standard natural language understanding (NLU) component. In such cases, a simple reply like ``Sorry, I don't know'' hurts the user's experience and limits the functionality of IPA. In this paper, we propose a two-stage shortlister-reranker recommender system to match third-party voice applications (skills) to unhandled utterances. In this approach, a skill shortlister is proposed to retrieve candidate skills from the skill catalog by calculating both lexical and semantic similarity between skills and user requests. We also illustrate how to build a new system by using observed data collected from a baseline rule-based system, and how the exposure biases can generate discrepancy between offline and human metrics. Lastly, we present two relabeling methods that can handle the incomplete ground truth, and mitigate exposure bias. We demonstrate the effectiveness of our proposed system through extensive offline experiments. Furthermore, we present online A/B testing results that show a significant boost on user experience satisfaction. 
\end{abstract}


%



\maketitle

\section{Introduction}
Intelligent personal assistants (IPA) such as Amazon Alexa, Google Assistant, Apple Siri and Microsoft Cortana that allow people to communicate with devices through voice are becoming a more and more important part of people's daily life. IPAs enable people to ask information for weather, maps, schedules, recipes and play games. The essential part of IPA is the Spoken Language Understanding (SLU) system which interprets user requests and matches voice applications (a.k.a skills) to it. SLU consists of both an automatic speech recognition (ASR) and a natural language understanding (NLU) component. ASR first converts the speech signal of a customer request (utterance) into text. NLU component thereafter assigns an appropriate domain for further response \cite{tur2011spoken}. 

However, utterance texts can be diverse and ambiguous, and sometimes contain spoken or ASR errors, which makes many utterances not able to be handled by the standard NLU system on a daily basis. As a result, they will trigger some NLU errors such as low confidence scores, unparsable, launch errors, etc. We call these utterances ``unhandled utterances''. IPAs typically respond to them by phrases such as ``Sorry, I don't understand''. However, these responses are not very satisfactory to the customers, and they harm the flow of the conversation. This paper focuses on developing a deep neural network based (DNN-based) recommender system (RS) to address this hard problem by recommending third-party skills to answer customers' unhandled requests. 

\begin{figure*}[h]
	\centering
	\includegraphics[width=0.8\textwidth]{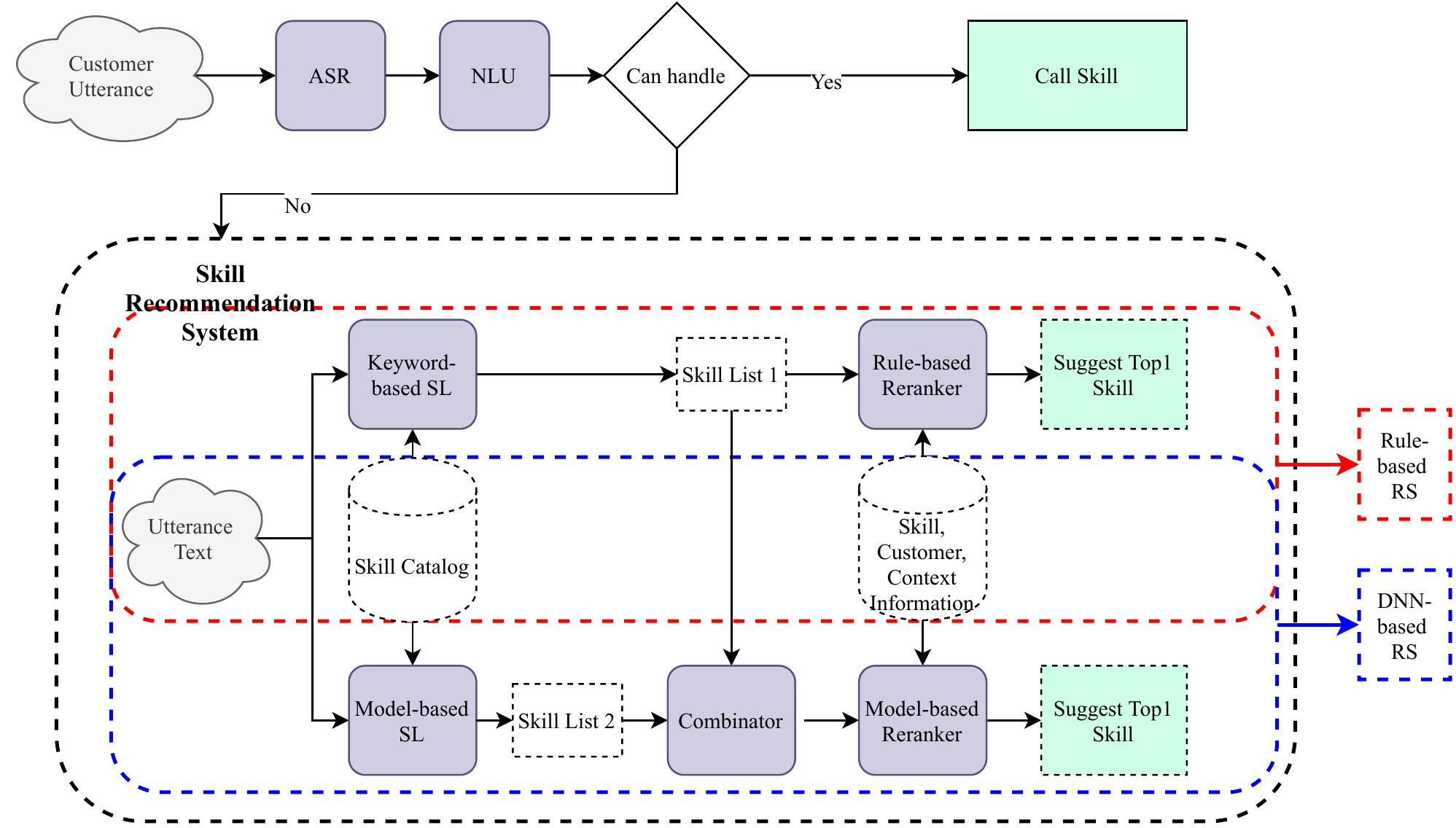}
	\caption{An overview of two stage skill recommender system.}
	\label{overview_system}
\end{figure*}

As our system utilizes a voice-based interface, only the top-1 skill is suggested to the customer. The whole process is illustrated in Figure \ref{overview_system}. The recommender system will first try to match a skill to the customer utterance, and if successful, the IPA responds with ``Sorry, I don't know that, but I do have a skill you might like. It's called $<$skill\_name$>$. Wanna try it? '' instead of simply saying ``Sorry, I don't know''. If customers respond ``Yes'', we call it a successful suggestion. Our goal is to improve both the customer accepted rate for the skill suggestion from the recommender system and the overall suggestion rate (percentage of unhandled utterances for which the RS suggests a skill).

We emphasize that building the above skill recommender system is not an easy task. One reason is that third-party skills are independently developed by third-party developers without a centralized ontology and many skills have overlapping capabilities. For example, to handle the utterance ``play the sound of thunder", skills such as ``rain and thunder sounds'', ``gentle thunder sounds", ``thunder sound", can all handle this request well. Another reason is that third-party skills are frequently added, and currently Alexa has more than one hundred thousand skills. Therefore, it is impossible to rely on human annotation to collect the ground truth labels for training.    

Before we launch our new DNN-based recommender system, we first build a rule-based recommender system to solve the ``skill suggestion task for unhandled utterance''. Rule-based system works as such: 1) when it is given a customer utterance, it invokes a keyword-based shortlister (Elasticsearch \cite{gormley2015elasticsearch}) to generate $K$ skill candidates; 2) a rule-based system picks one skill from the skill candidates list and suggests it to the customer for feedback; 3) If customer responds ``Yes", the system launches this skill. This is also the source where we collect our training data. One limitation for this automatically labeled dataset is that for a given utterance, we only collect the customer's response regarding a single skill. Thus, we have incomplete ground truth labels.

The rule-based system's shortlister only focuses on the lexical similarity between the customer utterance and the skill, which may omit good skill candidates. To remedy this limitation, we build a model-based shortlister which is able to capture the semantic similarity. We then combine both lists of skill candidates to form the final list. Our proposed DNN-based skill recommender system is composed of two stages, shortlisting and reranking. Shortlisting stage includes two components, shortlister and combinator. Reranking stage has the component reranker. The system works as follows. Given the customer utterance, model-based shortlister retrieves the top $K_1$ most relevant skills from the skill pool. These skills are combined with $K_2$ skills returned from the keyword-based shortlister of the rule-based RS in the combinator to form our final skill list. The reranker component ranks all skills in the final skill list. Based on the model score of the top-1 ranked skill, the skill recommender system decides whether to suggest this skill to the customer or not.

Biases are common in recommender systems as the collected data is observational rather than experimental. Exposure bias happens as users are only exposed to a part of specific items so that unobserved interactions do not always represent negative preference \cite{chen2020bias}. When we build our DNN-based recommender system, we find that exposure bias is a big obstacle. Specifically, we collect our training/testing data based on the rule-based system, and the rule-based exposure mechanism of this system degrades the quality of our collected data as positive labels are only received on skills suggested by the rule-based system. For example, for one utterance, we only have the customer's true response to one skill A, while it is highly likely that another more appropriate skill B also exists and we collect no positive customer response on skill B. A simple approach such as treating unobserved (utterance, skill) pairs as negative is problematic and hurts the model's performance as it is likely to mimic the rule-based system's decision to suggest skill A instead of skill B. We solve this by utilizing relabeling techniques, either collaborative-based or self-training, which is illustrated in Section \ref{subsec: relabeling}. Furthermore, we find that the exposure bias generates discrepancy between offline evaluation on test data and evaluation based on human annotation. In the end, we rely mainly on human annotation to draw the final conclusion. 

To sum up, the contribution of this work is threefold:
\begin{itemize}
	\item A new architecture is proposed to generate a skill shortlist by combining a lexical similarity focused keyword-based Shortlister (SL) and a semantic similarity focused model-based SL. We also propose a robust model-based SL with multi-task learning, which naturally incorporates meta information of skills into the prediction. The new model-based SL achieves better performance than the keyword-based SL based on human annotation metrics and offline metrics computed on test data. 
	\item Two relabeling approaches are proposed to solve the incomplete ground truth label and exposure bias problems. Both approaches significantly improve the reranker model's performance based on human annotation metrics.
	\item Recommender systems have widely changed people's daily life through many important applications. However, most of the works focus on developing complex architectures to better fit the observed data. When biases exist, this approach may not lead to better online metrics. We provide a concrete case study to demonstrate that exposure bias can lead to significant discrepancies between offline and online metrics.
\end{itemize}

\section{The Proposed Methodology}


\subsection{Two-stage architecture}
Our proposed architecture consists of two stages, \textit{shortlistin}g and \textit{reranking}. In the \textit{shortlisting} stage, for each utterance text ($u$), we call the shortlister module to get the top $K$ candidate skills $(S = \{s_1, s_2, \ldots, s_K\})$. The primary goal at this stage is to have a candidate skill list that has high recall to cover all relevant skills and low latency for computation. In the second \textit{reranking} stage, the reranker module assigns a relevancy score to each skill in the candidate skill list. Finally, we choose the skill with the highest relevancy score and compare this score to a pre-defined cutoff value. If it is larger than the cutoff value, we suggest this skill to the customer. Otherwise, the user is given the generic rejection sentence, e.g. ``Sorry, I don't know.''

\subsection{Shortlister}
We consider two types of shortlisters (SL): a keyword-based shortlister and a model-based shortlister. Both shortlisters can be formulated as follows. Assume the skill set (consists of skill\_ids) size is $N_s$. Given the input utterance $u$, SL computes a function $f^{\mathrm{SL}}(u, \theta)$, which returns a $N_s$ dimension score vector $O=(o_{1}, \ldots, o_{N_s})$. Each $o_k$ represents how likely skill $k$ is a good match to the utterance $u$. SL then returns the list of skill candidates with the top-K highest scores ordered by scores in descending order.    

In the keyword-based shortlister, we first index each skill using its keywords collected from various metadata (skill name, skill description, example phrases, etc), and then a search engine is called to find the most relevant $K$ skills to the utterance. We use Elasticsearch \cite{gormley2015elasticsearch} as our search engine as it is widely adopted in the industry and we find it to be both accurate and efficient. Specifically, Elasticsearch is called to measure the similarity score between each pair of utterance and skill by computing a matching score based on TF-IDF \cite{ramos2003using}. Top $K$ skills with the highest similarity scores are returned as the keyword-based shortlister list $S^{\mathrm{rule}}$.

In the model-based shortlister, we utilize a DNN-based model to compute the similarity scores. The model takes the utterance text $u$ as input, and $Y = (y_1, \ldots, y_{N_s^*})$ as the ground truth label, where $N^{*}_s$ is the skill set size that we used to train SL model and $y_k = 1$ if the $k$-th skill is suggested and accepted by the customer and $0$ otherwise. In our training data, the number of components of $Y$ that equals one is always one, where we exclude samples that customers provide negative feedback. As model-based SL's skill set only contains skills that exist in our training data, $N^{*}_s$ is significantly smaller than $N_s$ ($N^{*}_s$ is less than 10\% of $N_s$) which we use in the keyword-based shortlister..

\begin{figure*}[ht]
    \centering
	\begin{subfigure}{0.45\textwidth}
		\centering
		\includegraphics[height=6cm]{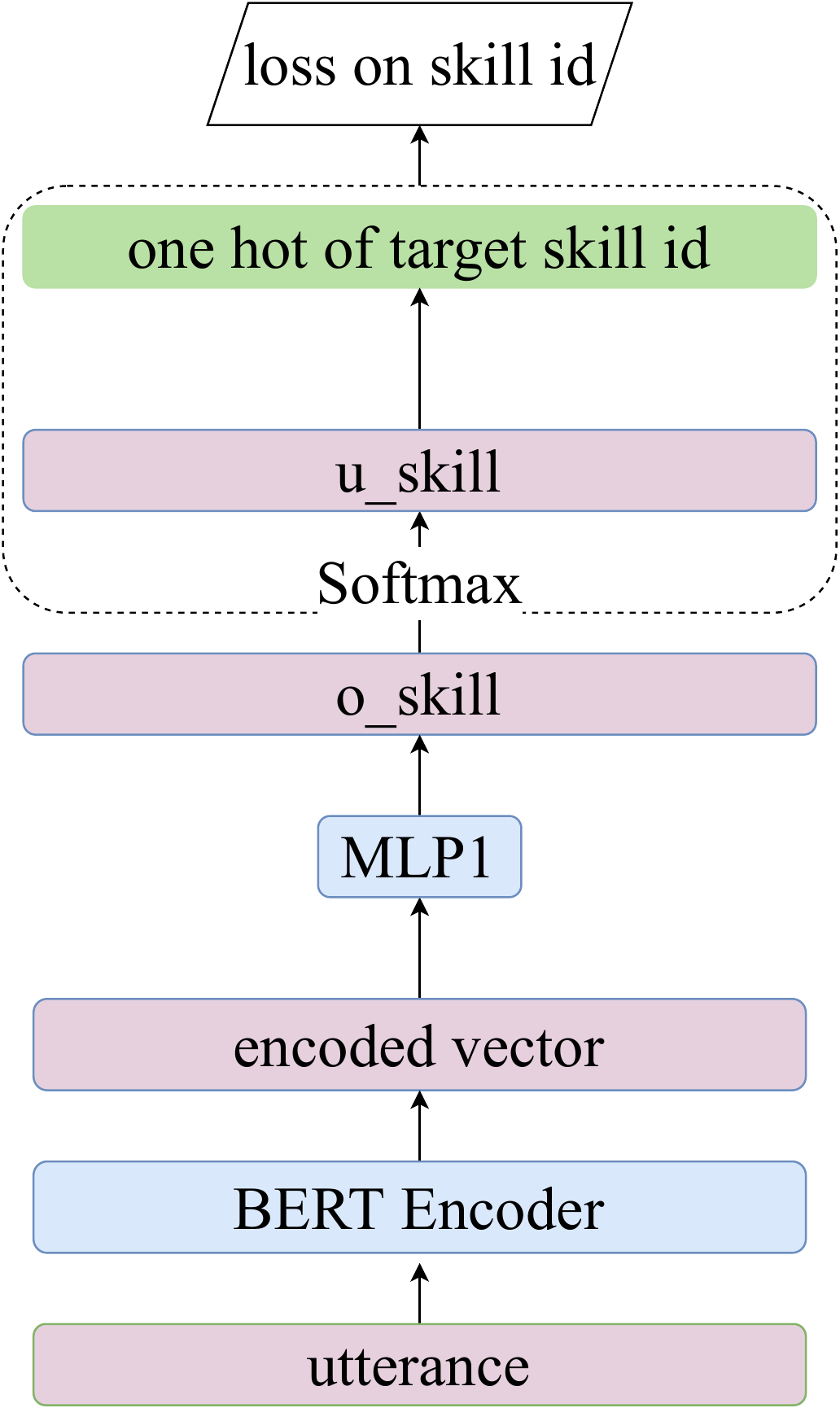}
		\caption{SL without multi-task learning}
		\label{fig: multi-task sl a}
	\end{subfigure}
	\begin{subfigure}{0.45\textwidth}
		\centering
		\includegraphics[height=6cm]{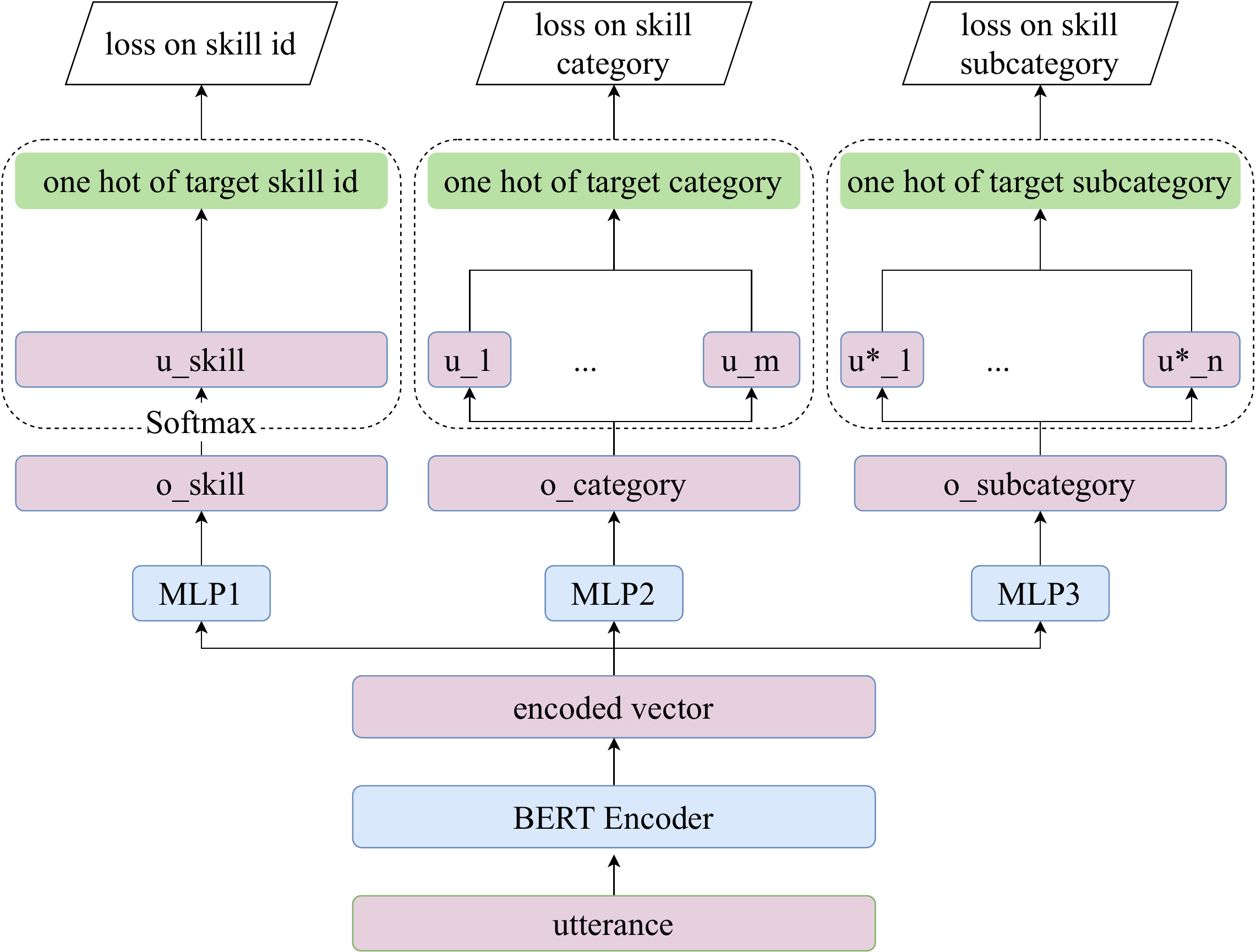}
		\caption{SL with multi-task learning}
		\label{fig: multi-task sl b}
	\end{subfigure}
	\caption{Model architecture of shortlister model}
	\label{multi-task sl}
\end{figure*}

Model-based shortlister works as follows. Utterance text $u$ is first fed to an encoder. Then, we feed the encoded vector to a two-layer multi-layer perceptron (MLP) of size $(128, 64)$ with activation function ``relu'' and dropout rate 0.2. The output is then multiplied by a matrix of size $N^{*}_s\times 64$ to compute $O = (o_{1}, \ldots, o_{N^{*}_s})$. For the encoder, we experiment with a RNN-based encoder, a CNN-based encoder and an in-house BERT \cite{devlin2018bert} encoder fine-tuned with Alexa data. We find that the BERT encoder has the best performance and we choose the first hidden vector of BERT output corresponding to [CLS] token as the encoded vector. In this paper, we only present the results with the BERT encoder. Please see Figure \ref{fig: multi-task sl a} for a demonstration. 

We experiment with two types of loss functions,
\begin{align}
	L_1 =& \sum_{k=1}^{N_s^{*}}-\{y_k \log \mathrm{sigmoid}(o_k)+ (1 - y_k)\log(1 - \mathrm{sigmoid}(o_k))\},  \label{eq:loss_sigmoid} \\
	L_2 =& \sum_{k=1}^{N_s^{*}}-y_k\log \mathrm{softmax}(O)_k, \label{eq: loss_multi_class}
\end{align}
where $\mathrm{softmax}(O)_k$ represents the $k$-th component of the vector $O$ after a softmax transformation. Here $L_1$ stands for one-versus-all logistic loss and $L_2$ is the multi-class logistic loss. In our experiment, we find that using different loss functions has little impact on the model's performance. In this paper, we show only results based on multi-class logistic loss.

Multi-task learning is a useful technique to boost model performance by optimizing multiple objectives simultaneously with shared layers. For our problem, skill category and subcategory are useful auxiliary information about the skill besides skill id as skill category/subcategory are tags assigned by its developers to skills based on their functionalities. Thus, in addition to multi-class logistic loss in equation \ref{eq: loss_multi_class} which only consider the skill id, we also experiment with a multi-task learning based SL model which minimizes the combined loss 
\begin{align*}
L = &w_1*\mathrm{loss(skill\_id)} + w_2*\mathrm{loss(skill\_category)} + \\ &w_3*\mathrm{loss(skill\_subcategory)},
\end{align*}
where the second and third loss functions have the same form as equation \ref{eq: loss_multi_class} and the ground truths are given by the skill category and subcategory. Here, we treat $(w_1, w_2, w_3)$ as hyper-parameters and the model architectures are illustrated in Figure \ref{fig: multi-task sl b}. In our experiments, we find that applying multi-task learning slightly improves the SL model's performance. Thus, we only report the results of models trained with multi-task learning in this paper. The selected model has $(w_1, w_2, w_3) = (1/3, 1/3, 1/3)$ based on a grid search. 

One limitation of the current model-based SL is that when a large number of new skills are added to the skill catalog, we need to update the SL model by retraining with the newly collected data from the updated skill catalog. A practical solution is to retrain the SL model every month.  

\subsection{Combinator}
\label{subsec: combinator}
In the DNN-based RS, unlike rule-based RS, we do not directly feed the skill candidates list ($S^{\mathrm{model}}$) returned from the shortlister component to the reranker. This is because the skill candidates list returned from model-based SL only contains skills that are collected in our training data which are suggested to customers based on the rule-based RS, and thus is incomplete and does not cover all important skills. Instead, we combine it with the skill candidate list returned from the keyword-based SL ($S^{\mathrm{rule}}$) by appending $S^{\mathrm{rule}}$ to it. We exclude all duplicate skills in the combination process, where the skills in $S^{\mathrm{rule}}$ which are also in $S^{\mathrm{model}}$ are removed. 

\subsection{Reranker}
\label{subsec: reranker}
\begin{figure}[h]
	\centering
	\begin{subfigure}{0.5\textwidth}
		\centering
		\includegraphics[width=8cm]{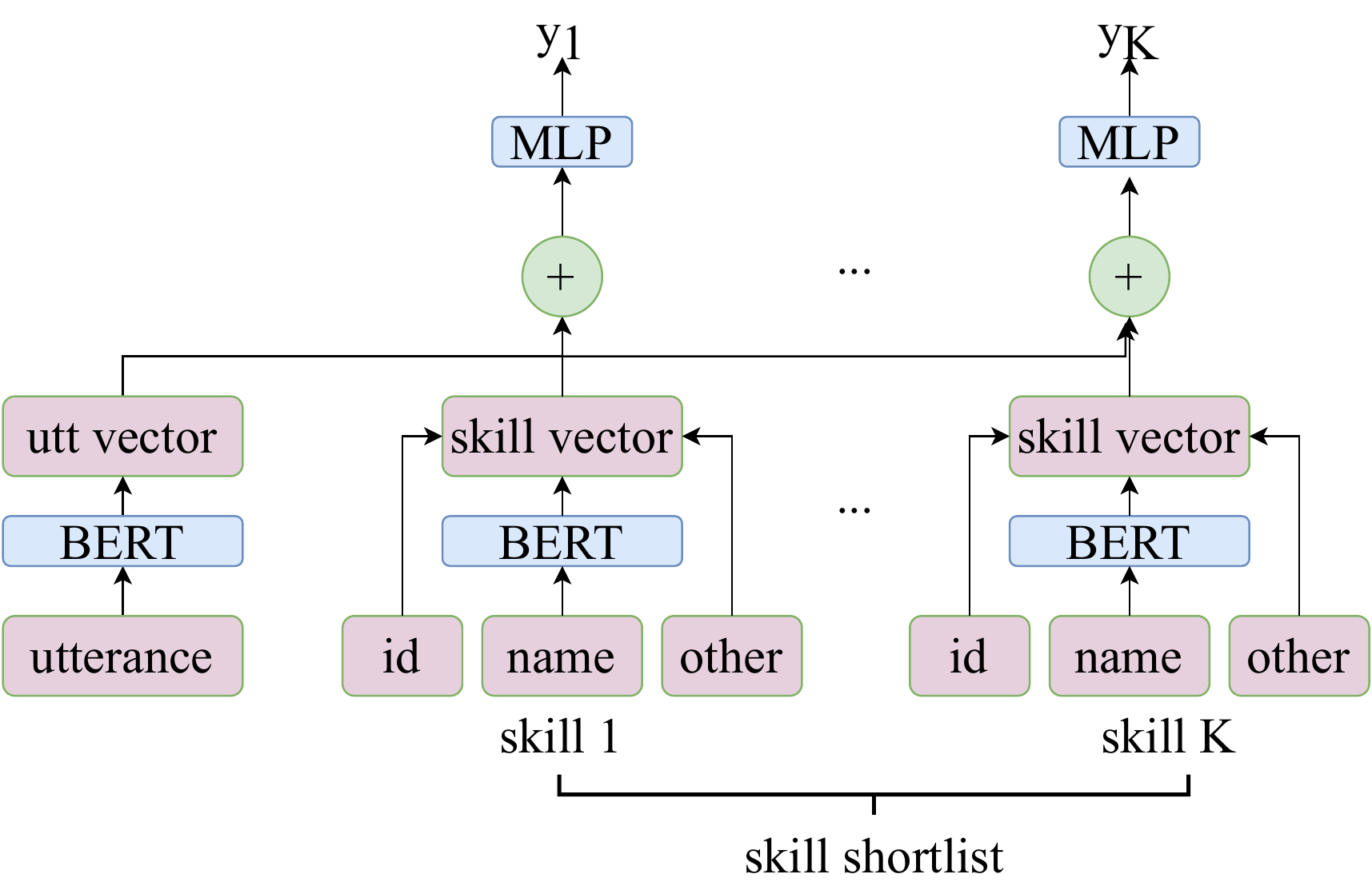}
		\caption{Pointwise Reranker}
		\label{fig: pointwise rr}
	\end{subfigure}
	\begin{subfigure}{0.5\textwidth}
		\centering
		\includegraphics[width=8cm]{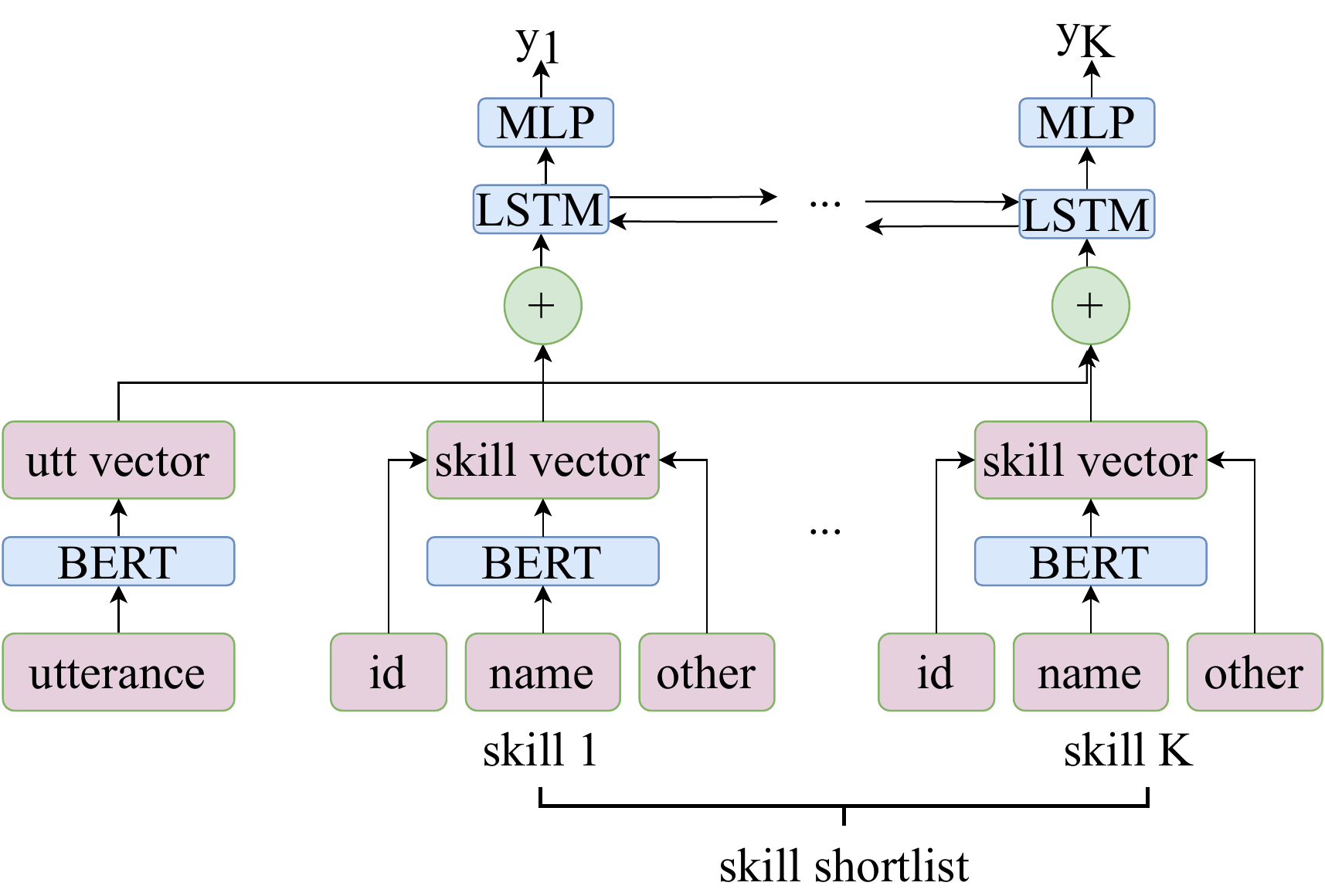}
		\caption{Listwise Reranker}
		\label{fig: listwise rr}
	\end{subfigure}
	\caption{Model architecture of reranker model}
	\label{fig: rr}
\end{figure}

The reranker model ranks the K skill candidate list returned from the shortlisting stage to produce a better ranking. We consider two types of models: a pointwise reranker and a listwise reranker. The architectures are shown in Figure \ref{fig: rr}. The model encodes both the utterance and the skill name with the same BERT-based encoder. Additionally, the skill encoder utilizes the following variables: skill id, skill score bin (three-level binned score returned from the shortlisting stage), skill category, skill popularity (0/1), and skill flag (a binary indicator of the skill returned from keyword-based SL or model-based SL). These variables are encoded via an embedding layer, then are
concatenated and fused (through a MLP layer) in the end to form the skill vectors. Utterance vector and skill vectors are concatenated and fed to a MLP layer to produce the predicted scores. Two architectures share the same loss function, which is the binary cross-entropy loss between target label $Y = (y_1, \ldots, y_{K})$ and predicted score $\hat{S} = (\hat{s}_1,\ldots,\hat{s}_K)$, i.e., 
\begin{equation*}
	L = \sum_{i=1}^{K} - y_i\log\hat{s}_i - (1 - y_i)\log(1 - \hat{s}_i).
\end{equation*}
The only difference between the listwise reranker and the pointwise reranker is that the former one has an additional bi-LSTM layer which makes the information flow freely between different skills. Thus, the final ranking of the listwise model considers all K skill candidates together. In our experiments, the listwise approach outperforms the pointwise one. 

We emphasize that the left tower of our architectures only utilizes the utterance. This part can be easily extended to incorporate user preference, session details or other contextual information to make more personalized recommendations. This is left for future exploration. 

\subsection{Relabeling}
\label{subsec: relabeling}
As pointed out in Section 1, our ground truth target $Y$ is incomplete: for each utterance, only one of the skills has a ground truth label based on customer feedback to the rule-based RS. Furthermore, as the distribution of suggested skills is determined by the rule-based RS, this adds exposure bias to our data. Our setting is close to the multi-label positive and unlabeled  (PU) learning \cite{yu2014large, kanehira2016multi, jain2017scalable, kim2020pseudo}with one major difference: our observed targets are not all positive and can be negative as well.

A naive way to solve the above incomplete label problem is to assign zeros (negatives) to all of the unobserved skills. However, this approach is not reliable: based on manual annotation, we find that frequently there are multiple  ``good'' skills that appear together in the skill candidate list. Assigning only one of them with a positive target confuses the model's label generation process and hurts the model's overall performance. Thus, we experiment with two relabeling approaches to alleviate this issue: collaborative relabeling and self-training relabeling. These two approaches borrow ideas from pseudo labeling \cite{lee2013pseudo} and self-training \cite{yarowsky1995unsupervised, rosenberg2005semi, ruder2018strong}, which commonly utilized in semi-supervised learning.     

\subsubsection{Collaborative relabeling}

Collaborative relabeling is a relabeling approach that borrows from kNN (k-nearest neighbors). For each target utterance, we first find all similar utterances in the training data and use the ground truth labels of these neighbors to relabel the target utterance. Specifically, for each utterance, we first compute its embedding based on a separate pre-trained BERT encoder. Then, for each target utterance, we compute its correlation to all of the other utterances based on cosine similarity and keep only the top $m$ pairs with correlation at least $r$. We then combined the target information from these filtered neighbors and get a list of tuples $\{(\mathrm{skill}_1, p_1, n_1), \ldots, (\mathrm{skill}_k, p_k, n_k)\}$, where $(\mathrm{skill}_i, p_i, n_i)$ represents that there are $n_i$ neighbors with suggested skill $\mathrm{skill}_i$ and average accept rate $p_i$. We then filter out all skills with $n_i$ smaller than $n_c$ and $p_i$ smaller than $p_c$. For the remaining skills, if they appear in the target utterance's shortlisting list and have missing labels, we label them as positive (negative) examples with probability $p_i$ ($1-p_i$). Here $n$, $r$, $n_c$, $p_c$ are hyperparameters and we find the optimal choice through grid search. From our experiment, $m=100$, $r=0.995$, $n_c=6$, $p_c=0.45$ achieves the best performance on the validation dataset.

\subsubsection{Self-training relabeling}

Self-training relabeling is a relabeling method that uses the model's prediction to relabel the ground truth target. The algorithm is summarized in Algorithm \ref{algorithm: self-training}. We experiment with a constant threshold ($c_i = c$) and an adaptive threshold where we increase the threshold slowly over the iterations, that is $c_i = c + 0.1*i$. We find that the adaptive threshold with increasing cutoff value across iterations works better. As we increase the iterate $i$, our training data contains more and more positive labels due to relabeling, and we need to increase the threshold simultaneously to avoid adding false positive labels. The optimal iteration number $i^*$ and the optimal threshold are selected by a hyper-parameter search that minimizes that loss on validation data. Based on our experiment, $i^*=5$, $c=0.3$ works the best.  
\begin{algorithm}[ht]
	\SetAlgoLined
	\textbf{Initialization}: Let $i = 0$, $N=10$. Set current model as the baseline reranker model\;
	\While{$i < N$}{
		Run current model on the training data to get predicted scores\;
		Relabel all skills in the skill shortlist with a predicted score above a cutoff value ($c_i$) to 1. We do not overwrite the skill with observed customer feedback\;
		Update the current model by retraining the model with relabeled training data.
	}
	\caption{Self-training relabeling}
	\label{algorithm: self-training}
\end{algorithm}

\section{Experiments}
We collect two months' data (2020/4 and 2020/7) from Alexa traffic (unhandled English utterances only) of devices in the United State under the rule-based recommender system as our dataset. Individual users are de-identified in this dataset. The last week of the dataset is used as testing and the second to last week's data is used as validation. The rest is used for training. The models are all trained with AWS EC2 p3.2xlarge instance. 

Using solely this test data to evaluate model performance results in severe exposure bias due to the aforementioned reasons of partial observation. More specifically, a matched skill can have ground truth label 0 only because this skill is not suggested to the customer by the rule-based RS. Thus, we randomly sample 1,300 utterances from our test dataset to form our human annotation dataset. We combine the predictions on this dataset from all of our models (including the various shortlisters) and the binary labels are manually annotated by human annotators. 

\subsection{Shortlister Model Comparison}
We experiment with two different sizes of skill set for the model-based SL model, where the former vocabulary set contains the top 2,000 most frequently observed skills in the training dataset ($N^{*}_s=2,000$) and the latter one contains all skills that are observed at least 2 times ($N^{*}_s=11,000$) .

Table \ref{tab:sl_result} summarizes shortlister models' performance. Due to Alexa's critical data confidential policy, we are not allowed to directly report their absolute metric scores. Instead, we report the normalized relative difference of each method when compared to the baseline method ``keyword-based SL''. We present two common metrics in information retrieval (Precision@K and NDCG@K) to evaluate the models. Recall metrics are not provided as they are technically impossible to compute: there is more than one relevant skill for most utterances and we do not have access to this ground truth. From Table \ref{tab:sl_result}, we see that the model-based SL outperforms keyword-based SL in terms of both human annotation metrics and offline metrics computed on test data. In test data, the positive skill is derived from the rule-based RS and is always in the skill candidate list (length = 40) generated by the keyword-based SL. Thus, Precision@40 of keyword-based SL has the highest possible value when computed on test data, which is larger than model-based SL. However, this does not prove that keyword-based SL is better. Furthermore, we find that using a large skill set size ($N_s^*=11,000$) improves the SL model's performance. Thus, we use SL with $N_s^*=11,000$ in the two-stage RS comparison.

\begin{table*}[h]
	\centering
	\renewcommand{\tabcolsep}{2pt}
	\begin{tabular}{lrrrrrrr} 
		\toprule
		\multirow{2}{*}{\textbf{Method}} & \multirow{2}{*}{$\mathbf{N_s^*}$} & \multicolumn{2}{c}{\textbf{Human Annotation Metric}}  & \multicolumn{4}{c}{\textbf{Metric computed on test data}} \\
		\cmidrule(lr){3-4} \cmidrule(lr){5-8}
		& & \textbf{Pre@1} & \textbf{Pre@3} & \textbf{Pre@5} & \textbf{NDCG@5} & \textbf{Pre@40} & \textbf{NDCG@40}\\\midrule
		keyword-based SL    &       & +0.00\% & +0.00\% & +0.00\% & +0.00\%	& \textbf{+0.00\%} &	+0.00\% \\
		model-based SL & 2,000 & N/A  & N/A & +103.46\% & +158.79\%	& -7.20\% &	+80.09\% \\
		model-based SL & 11,000 &  \textbf{+162.68\%} &  \textbf{+171.72\%} & \textbf{+108.29\%} & \textbf{+164.54\%}	& -4.60\% & \textbf{+84.25\%} \\
		 \bottomrule
	\end{tabular}
	\caption{Summarization of shortlister models' performance. Normalized relative difference of each method when compared to baseline method ``keyword-based SL'' is presented. Positive values (+) implies that the method outperforms baseline method.}
	\label{tab:sl_result}
\end{table*} 

\subsection{Two-stage recommender system Comparison}
We considered the following four reranker models, and compared their performance by reranking on the skill shortlist obtained from keyword-based SL and model-based SL (for model-based SL, we use the combined skill shortlist as illustrated in Section \ref{subsec: combinator}), respectively. 

\begin{itemize}
	\item \textbf{Pointwise:}  reranker model with pointwise architecture as introduced in Section \ref{subsec: reranker}.
	\item \textbf{Listwise:}  reranker model with listwise architecture as introduced in Section \ref{subsec: reranker}.
	\item \textbf{Collaborative:}  reranker model with listwise architecture and trained with collaborative relabeling (Section \ref{subsec: relabeling}).
	\item \textbf{Self-training:}  reranker model with listwise architecture and trained with self-training relabeling (Section \ref{subsec: relabeling}).	
\end{itemize} 
Table \ref{tab: rs_result} summarizes the two-stage recommender systems' performance. As in the previous Section, we only report the normalized relative difference of each method when compared to the baseline method ``Listwise + keyword-based SL''. We present precision, recall, F1-score of the model at cutoff point 0.5, and the precision of the model at different suggestion rates (25\%, 40\%, 50\%, 75\%) as our metrics. Here we control the model's suggestion rate by changing the cutoff value. For example, if we want a higher suggestion rate, we decrease the cutoff value and vice versa.

\begin{table*}[h]
	\centering
\resizebox{\textwidth}{!}{%
	\renewcommand{\tabcolsep}{2pt}
	\begin{tabular}{lrrrrrrrrrr} 
		\toprule
		\multirow{2}{*}{\textbf{Method}} & \multicolumn{7}{c}{\textbf{Human Annotation Metric}} & \multicolumn{3}{c}{\textbf{Metric computed on test data}}  \\
		\cmidrule(lr){2-8} \cmidrule(lr){9-11}
		& \textbf{Pre@25\%} & \textbf{Pre@40\%}  & \textbf{Pre@50\%}  & \textbf{Pre@75\%} & \textbf{Precision} & \textbf{Recall} & \textbf{F1} & \textbf{Precision} & \textbf{Recall} & \textbf{F1}\\\midrule
Pointwise + keyword-based SL     & -10.19\% & -6.51\% & -6.73\% & -9.93\% & -10.51\% & -63.11\% & -58.69\%                                                                 & -7.08\% & -63.76\% & -57.12\%  \\
Listwise + keyword-based SL      &	  +0.00\% &  +0.00\% &  +0.00\% &  +0.00\% &  +0.00\% &   +0.00\% &   +0.00\%                                                         &  +0.00\% &	+0.00\%	& +0.00\%   \\
Collaborative + keyword-based SL  &   +3.70\% &  +4.11\% &  +2.05\% & -0.69\% &   +0.74\% &  +16.39\% &  +13.19\%                                                        &  -15.05\% & +2.86\% & -2.37\% \\
Self-training + keyword-based SL &	  +8.33\% &  +9.59\% &  +2.05\% & -1.85\% & -10.69\% & +108.20\% &  +67.55\%                                                       &  -25.94\% & +87.06\% &	+34.83\% \\
Listwise + model-based SL     &	  +0.46\% &  +7.53\% &  +3.51\% &  +5.31\% &  -5.10\% &  -5.74\% &  -5.62\%                                                             & \textbf{+2.90\%} & +10.19\% & +8.25\% \\
Collaborative + model-based SL &	 \textbf{+10.19\%} &  +9.59\% &  \textbf{+6.73\%} &  \textbf{+6.47\%} &   \textbf{+1.55\%} &  +22.13\% &  +17.78\% & -15.88\% &	+10.48\% & +2.34\% \\
Self-training + model-based SL&	  +7.87\% & \textbf{+11.99\%} &  +5.26\% &  +4.39\% & -19.89\% & \textbf{+158.20\%} &  \textbf{+83.75\%}              & -31.97\% & \textbf{+128.07\%} & \textbf{+42.77\%} \\
		\bottomrule
	\end{tabular}
}
	\caption{Summarization of two-stage recommender systems' performance. Normalized relative difference of each method when compared to baseline method ``Listwise + keyword-based SL'' is presented. Positive values (+) implies that the method outperforms baseline method.}
	\label{tab: rs_result}
\end{table*} 

\begin{figure*}[ht]
	\centering
	\begin{subfigure}{0.45\textwidth}
		\centering
		\includegraphics[width=1\textwidth]{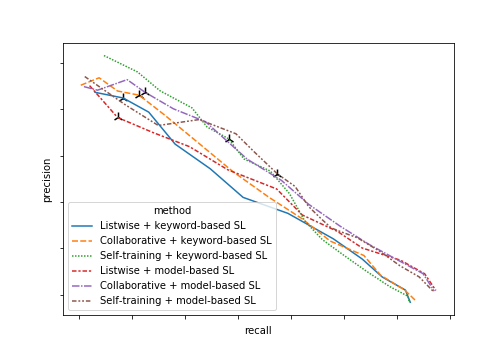}
		\caption{Precision-recall curve computed on human annotation data}
		\label{fig: pr_annotate}
	\end{subfigure}
	\begin{subfigure}{0.45\textwidth}
		\centering
		\includegraphics[width=1\textwidth]{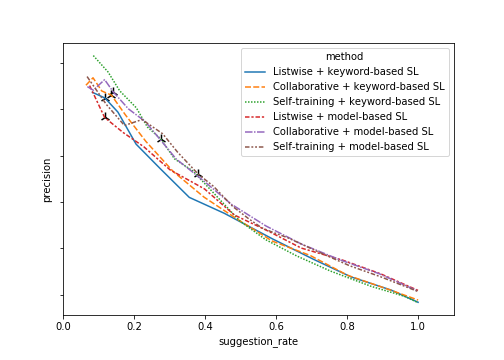}
		\caption{Precision v.s. suggestion rate computed on human annotation data}
		\label{fig: pr_suggestion_rate_offline}
	\end{subfigure}
	
	\begin{subfigure}{0.45\textwidth}
		\centering
		\includegraphics[width=1\textwidth]{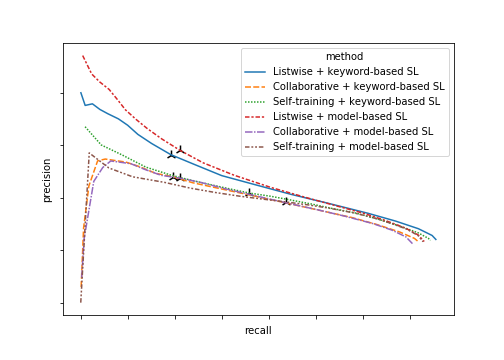}
		\caption{Precision-recall curve computed on test data}
		\label{fig: pr_offline}
	\end{subfigure}	
	\begin{subfigure}{0.45\textwidth}
		\centering
		\includegraphics[width=1\textwidth]{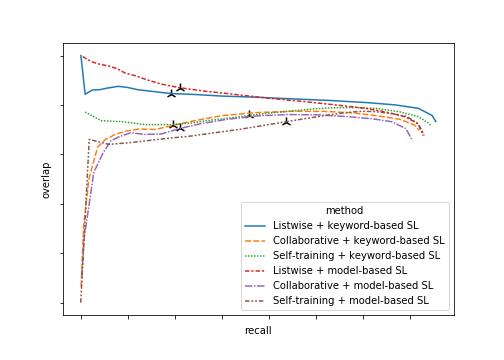}
		\caption{Overlap v.s. recall computed on test data.}
		\label{fig: overlap_offline}
	\end{subfigure}
	\caption{Model performance of reranker model. The model's metrics with cutoff point 0.5 is masked.}
	\label{fig: pre_recall_two_stage}
\end{figure*}

From Table \ref{tab: rs_result}, we find that it is hard to compare models based on precision, recall and F1-score as different models have very different recall levels. Thus, we also draw their precision-recall curves in Figure \ref{fig: pre_recall_two_stage}. From these figures, we find that there is a significant mismatch between human annotation metrics and metrics computed with offline test data. For example, in human annotation metrics, both collaborative and self-training relabeling improve the model performance. However, the opposite trend is observed on metrics computed on test data. In Figure \ref{fig: overlap_offline}, we plot the curve of overlap (the probability that the model suggests the same skill as rule-based RS) v.s. recall. We discover that metrics computed on test data tend to overestimate a model's performance if its overlap with rule-based RS is high. This is intuitively reasonable as all positive ground truth labels are observational and can only be found in skills suggested with rule-based RS. This mismatch on metrics is due to exposure bias. Other works in the literature also find similar patterns and conclude that conventional metrics computed based on observation data suffer from exposure bias and may not be an accurate evaluation of the recommender system's performance \cite{schnabel2016recommendations, yang2018unbiased, chen2020bias}. In our experiment, we use human annotation metrics to do a fair comparison between different models.  

We find that both collaborative and self-training relabeling improves the model's performance, and reranker models using skill list from model-based SL (combined list) outperform those that use skill list from keyword-based SL. This also justifies using model-based SL, as opposed to keyword-based SL. We also find that listwise reranker architecture significantly outperforms the pointwise reranker architecture.  The overall winner is Collaborative + model-based SL. 

For inference in production, we utilize AWS EC2 c5.xlarge instance and the 90\% quantile of total latency of model-based RS is less than 300ms.

\subsection{Sensitivity Analysis}
In the shortlisting stage, both keyword-based SL and model-based SL firstly returns a skill candidate list of length 40. Then, in model-based SL, its skill candidate list is combined with the keyword-based SL's list to form a combined list that is fed to the reranker model. Based on human annotation, we find that the most relevant skills are often returned in the top 10 candidates of the model-based SL's candidate list. In this section, we analyze whether reducing the candidate list's length of the model-based SL from 40 to 10 affects the overall RS performance. If the difference is not significant, one can instead rely on the top 10 candidates from model-based SL and enjoy faster inference during runtime.

\begin{figure}[ht]
	\centering
	\includegraphics[width=0.45\textwidth]{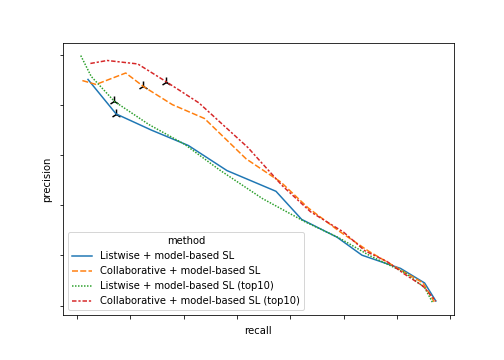}
	\caption{Precision-recall curve of DNN-based RS's performance with skill candidate length 40 v.s. 10 computed on human annotation data.}
	\label{fig: pr_40_vs_10}
\end{figure}

Comparison of DNN-based RS's performance with skill candidate length 40 v.s. 10 is provided in Figure \ref{fig: pr_40_vs_10}. We find that both approaches have roughly the same performance. The collaborative relabeling with skill candidate length 40 (yellow line) seems to be worse than that with skill candidate length 10 (red line) when recall is low. However, this is mainly due to the variation as only a small-sized human annotation dataset is available for the evaluation when recall level is low. 

\subsection{Ablation Study}
In this section, we study the contribution of each feature of the skills to the reranker model's performance. We choose the best model "Collaborative + model-based SL" as our baseline, and remove features one at a time while keeping all other features. Table \ref{tab: ablation} shows the result. We find that features like skill id, skill name and skill score bin are the most important and removing them has a big negative impact on the model's performance.

\begin{table}[h]
	\centering
		\renewcommand{\tabcolsep}{2pt}
		\begin{tabular}{lrrr} 
			\toprule
			\multirow{2}{*}{\textbf{Method}} & \multicolumn{3}{c}{\textbf{Human Annotation Metric}}\\
			\cmidrule(lr){2-4} 
			& \textbf{Pre@25\%} & \textbf{Pre@50\%}  & \textbf{Pre@75\%} \\\midrule
			\sout{skill id} & -2.52\%	& -1.88\% &	-2.17\%\\
			\sout{skill name} & -1.26\%	& -0.94\%	& -2.60\%\\
			\sout{skill score bin} & -2.94\%	& -2.81\%	& -1.30\%\\
			\sout{skill category} & +1.68\%	& +2.19\%	& -1.08\%\\
			\sout{skill popularity} & -0.42\%	& +0.31\%	& -0.43\%\\
			\sout{skill flag} & -0.42\%	& +1.56\%	& -0.87\%\\
			\bottomrule
		\end{tabular}
	\caption{Summarization of ablation study. It reports normalized relative difference when removing each feature from baseline model.}
	\label{tab: ablation}
\end{table} 

\section{Online Experiment}

We compare our DNN-based RS with rule-based RS through online A/B testing after observing the improvement in the offline metrics. We find that the new  DNN-based RS significantly increases the average accept rate by 1.65\% and reduces both the overall friction rate of customers and the customer interruption rate by 0.41\% and 3.39\%, respectively. The new DNN-based RS also suggests more diverse skills to customers: with the new model, customers discover and enable more skills. The increase of average number of enabled skills per customer can also improve the engagement of the users to Alexa in the long run. From the A/B testing, we find that the number of days a customer interacted with at least one skill has increased by 0.11\% with DNN-based RS. 

\section{Related Work}
Recommender system is the last line of defense to filter overloaded information and suggest items that users might like to them proactively. Recommender systems are mainly categorized into three types: content-based, collaborative filtering and a hybrid of both. Content-based RS recommends based on user and item features. They are most suitable to handle cold-start problems, where new items without user-item interaction data need to be recommended. Collaborative filtering \cite{sarwar2001item, linden2003amazon}, on the other hand, recommends by learning from user-item past interaction history through either explicit feedback (user's rating, etc) or implicit feedback (user's click history, etc). Hybrid recommender systems integrate two or more recommendation techniques to gain better performance with fewer drawbacks of any single technique \cite{burke2002hybrid}. \cite{burke2002hybrid, zhang2019deep} provide thorough reviews of recommender systems. Traditional recommender techniques include matrix factorization \cite{he2016fast}, factorization machine \cite{rendle2010factorization}, etc. In recent years, deep learning techniques are integrated with recommender systems to better utilize the inherent structure of the features and to train the system end-to-end. Some important works in this realm include NeuralCF \cite{he2017neural}, DeepFM \cite{guo2017deepfm}, Wide\&Deep model \cite{cheng2016wide} and DIEN \cite{zhou2019deep}. Deep learning based recommender systems gain great success in industry as well. For example,  \cite{covington2016deep} proposed a two-stage recommender system for youtube. The system is separated into a deep candidate generation model and a deep ranking model. Some other notable works include \cite{naumov2019deep, grbovic2018real, zhou2018deep, zhou2019deep}.

In our work, collecting ground truth labels based on human annotation is impossible due to the large volume of skills. Therefore, we rely on observation data collected from a rule-based system to train our model. This adds exposure bias to the problem as the rule-based system controls which skill is suggested to the users and hence the collected labels. Such exposure biases generate discrepancy between offline and online metrics \cite{schnabel2016recommendations, yang2018unbiased, chen2020bias}. Some previous works try to solve this issue using propensity score \cite{yang2018unbiased} in evaluation or sampling \cite{chen2019samwalker, ding2019reinforced} in training.      

Our work is also highly related to domain classification in SLU. Domain classification is an important component in standard NLU for intelligent personal assistants. They are usually formulated as a multi-class classification problem. Traditional NLU component usually covers tens of domains with a shared schema, but it can be extended to cover thousands of domains (skills) \cite{kim2018scalable}. Contextual domain classification using recurrent neural network is proposed in \cite{xu2014contextual}. \cite{chen2016end} studies an improved end-to-end memory network.  \cite{kim2018scalable} proposes a two-stage shortlister-reranker model for large-scale domain classification in a setup with 1500 domains with overlapped capacity. \cite{kim2020pseudo} proposes to use pseudo labeling and negative system feedback to enhance the ground truth labels.

\section{Conclusion}
In this paper, we propose a two-stage shortlister-reranker based recommender system to match skills (voice apps) to handle unhandled utterances for intelligent personal assistants. We demonstrate that by combining candidate lists returned from a keyword-based SL and a model-based SL, the system generates a better skill list that covers both lexical similarity and semantic similarity. We describe how to build a new system by using observed data collected from a baseline rule-based system, and how the exposure biases generate discrepancy between offline and human metrics. We also propose two relabeling methods to handle the incomplete ground truth target issue. Extensive experiments demonstrate the effectiveness of our proposed system.

\section{Acknowledgments}
We thank Emre Barut, Melanie Rubino and Andrew Arnold for their valuable feedback on this paper.

\bibliographystyle{ACM-Reference-Format}
\bibliography{reference}

\end{document}